\documentclass[]{ceurart}
\usepackage{listings}
\usepackage{graphicx}
\usepackage{subcaption}

\begin{document}

\copyrightyear{2024}
\copyrightclause{Copyright for this paper by its authors.
  Use permitted under Creative Commons License Attribution 4.0
  International (CC BY 4.0).}

\conference{CLEF 2024: Conference and Labs of the Evaluation Forum, September 9-12, 2024, Grenoble, France}

\title{Fine-Grained Classification for Poisonous Fungi Identification with Transfer Learning}

\title[mode=sub]{Notebook for the LifeCLEF Lab at CLEF 2024}

\author[1]{Christopher Chiu}[%
orcid=0000-0002-4219-1795,
email=cchiu65@gatech.edu,
]
\cormark[1]
\fnmark[1]

\author[1]{Maximilian Heil}[%
orcid=0009-0002-6459-6459,
email=mheil7@gatech.edu,
]
\cormark[1]
\fnmark[1]

\author[1]{Teresa Kim}[%
orcid=0009-0002-4514-3710,
email=tkim654@gatech.edu,
]
\cormark[1]
\fnmark[1]

\author[1]{Anthony Miyaguchi}[%
orcid=0000-0002-9165-8718,
email=acmiyaguchi@gatech.edu,
]
\cormark[1]
\fnmark[1]
\address[1]{Georgia Institute of Technology, North Ave NW, Atlanta, GA 30332, USA}

\cortext[1]{Corresponding author.}
\fntext[1]{These authors contributed equally.}

\begin{abstract}
  FungiCLEF 2024 addresses the fine-grained visual categorization (FGVC) of fungi species, with a focus on identifying poisonous species. This task is challenging due to the size and class imbalance of the dataset, subtle inter-class variations, and significant intra-class variability amongst samples. In this paper, we document our approach in tackling this challenge through the use of ensemble classifier heads on pre-computed image embeddings. Our team (DS@GT) demonstrate that state-of-the-art self-supervised vision models can be utilized as robust feature extractors for downstream application of computer vision tasks without the need for task-specific fine-tuning on the vision backbone. Our approach achieved the best Track 3 score (0.345), accuracy (78.4\%) and macro-F1 (0.577) on the private test set in post competition evaluation. Our code is available at \url{https://github.com/dsgt-kaggle-clef/fungiclef-2024}.
\end{abstract}

\begin{keywords}
  Fine-Grained Visual Categorization (FGVC) \sep
  Poisonous Fungi Identification \sep
  Transfer Learning \sep
  Vision Transformers \sep
  CEUR-WS
\end{keywords}

\maketitle

\section{Introduction}

Classifying a fungi’s species and toxicity accurately and efficiently goes beyond simple image classification; it requires discerning subtle differences between species within the intricate context of fine-grained visual categorization (FGVC). FGVC is more challenging than regular image classification tasks due to small inter-class variations - high similarities between genetically related fungi, and high intra-class variances at the same time as the image observation depends on several factors, such as genotype, age, time of year, and local conditions. This paper is part of FungiCLEF 2024 \cite{fungiclef2024} competition, part of the LifeCLEF \cite{lifeclef2024} lab series.

\begin{figure}[!h]
  \centering
  \begin{subfigure}[b]{0.3\textwidth}
    \includegraphics[width=\columnwidth, height=100pt]{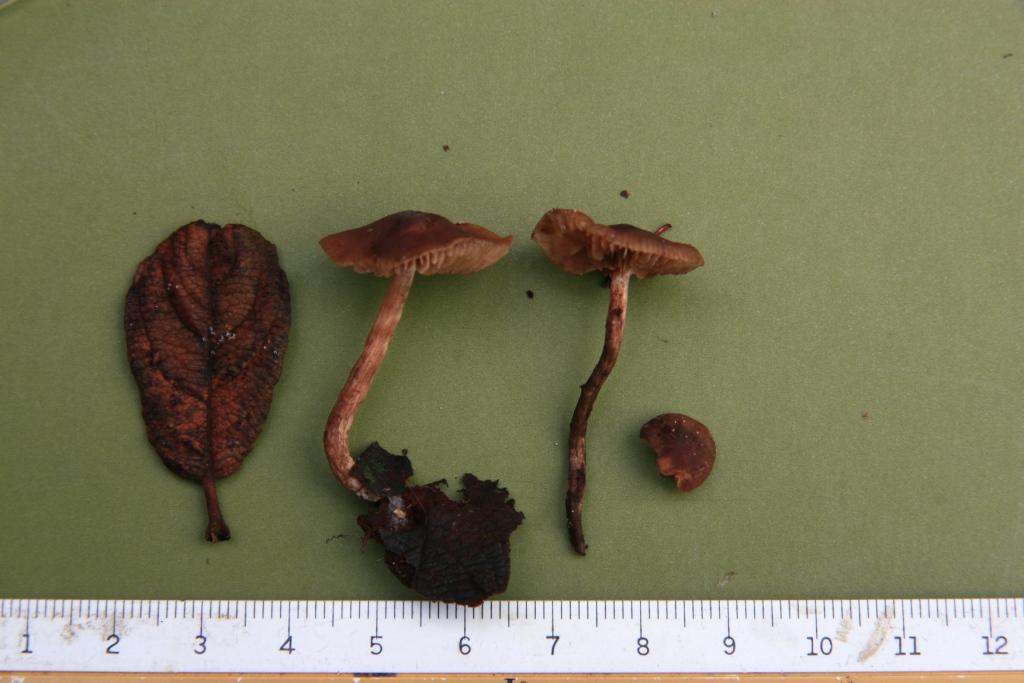}
  \end{subfigure}
  \hfill 
  \begin{subfigure}[b]{0.3\textwidth}
    \includegraphics[width=\columnwidth, height=100pt]{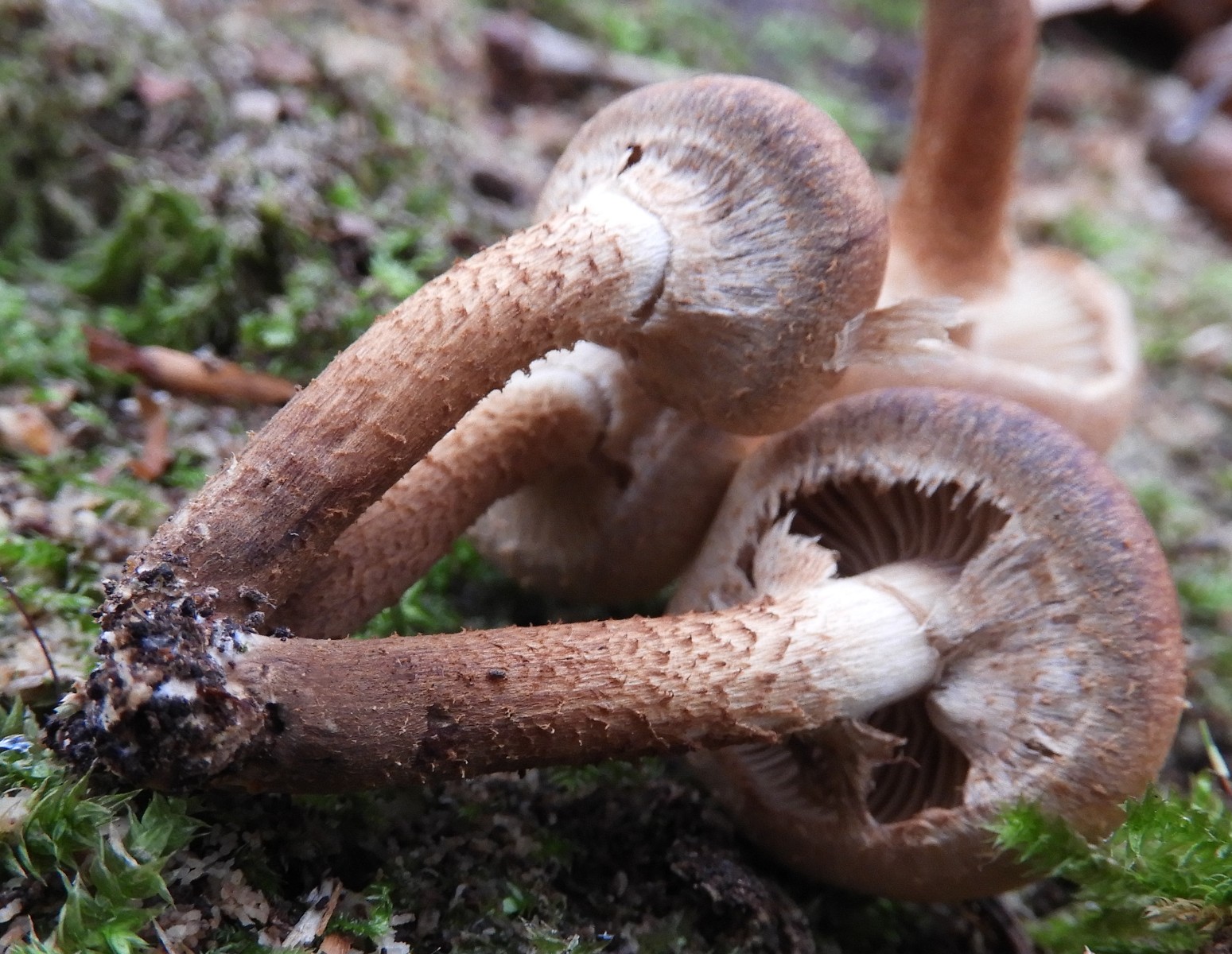}
  \end{subfigure}
  \hfill 
  \begin{subfigure}[b]{0.3\textwidth}
    \includegraphics[width=\columnwidth, height=100pt]{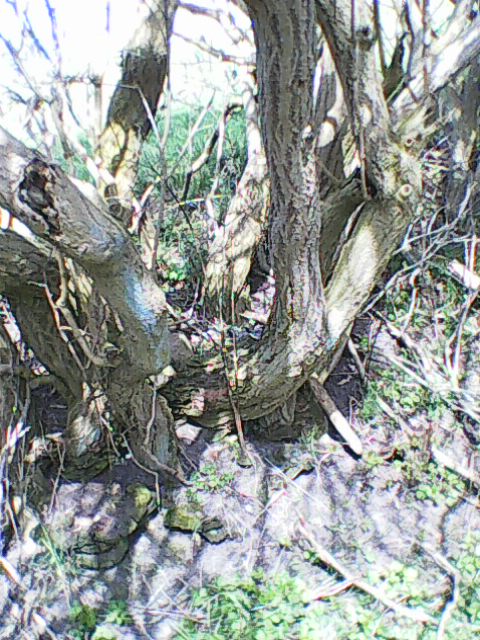}
  \end{subfigure}
  \caption{Sample images from the dataset. The images of fungi have high degree of variability across lighting, substrate, focus, subject, and other image features. This poses additional challenge in effectively training a classifier model.}
  \label{fig:overview-fungi}
\end{figure}

\subsection{Dataset Overview}

The featured dataset for the FungiCLEF competition \cite{fungiclef2024} is the Danish Fungi dataset \cite{Picek_2022_WACV}. This dataset includes a training set (DF20), which includes 356,770 images over 1,604 different classes of fungi, and a validation / testing dataset (DF21), consisting of 60,832 images over 2,713 species of fungi, covering a year's worth of observation. For species within the validation dataset that were not in the training dataset, they were marked as an "unknown" class.
The dataset provides both full sized images (110GB) and downsized images (300px max dimension, 5.6GB). It also provides metadata for the fungi images including date, location, substrate and metasubstrate of the fungi growth, and the full taxonomical ranks of the classified fungi species, including phylum, class, order, family, and genus. 

\begin{figure*}
  \centering
  \includegraphics[width=0.8\linewidth]{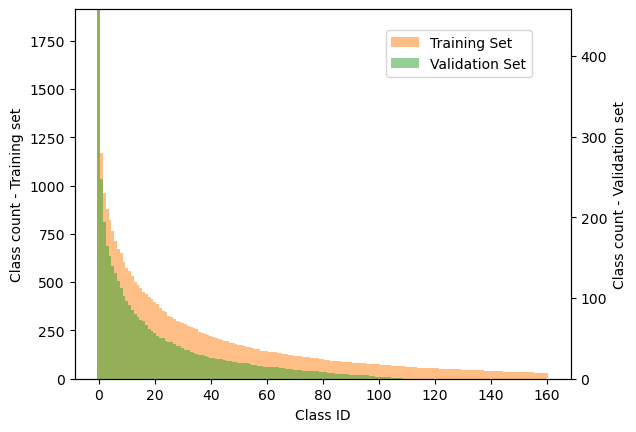}
  \caption{Distribution of classes in DF20 (training set) and DF21 (validation set). We observe that the two datasets have different class distributions. Both datasets have significant class imbalance, as signified by the long tail in classes with low counts.}
  \label{fig:class-distribution}
\end{figure*}

These two datasets do not have the same distribution of classes (Figure \ref{fig:class-distribution}). Moreover, there was significant class imbalance in both datasets, with the most common class having 1,913 images, and the least common class only \textasciitilde 30 images in DF20, and down to only one observation for some species in DF21. There were also significant variations in terms of lighting, background, and clarity, due to the real-world conditions under which fungi were photographed (Figure \ref{fig:overview-fungi}). This adds another level of complexity on this task - Fungi classes are not only hard to distinguish due to inter-class variations or intra-class variance, but also due to varying image quality and image features. This highlights the need for a robust model in effectively performing fine-grained classification on this rich and varied dataset.  

\subsection{Related Work}

State-of-the-art work on this dataset primarily utilizes models such as Swin Transformer \cite{SwinTransformer} and MetaFormer \cite{MetaFormer}. 
However, results from FungiCLEF 2023 \cite{FungiCLEF2023Comp} underscore limitations in current research, where the best accuracy from participants have not improved significantly since the competition's inception in 2022 \cite{FungiCLEF2022}. 
Last year’s winner incorporated metadata into the model with MetaFormer as the vision model \cite{2023winner}, and utilized Seesaw Loss \cite{SeesawLoss} to handle class imbalance. 
This led to a macro F1 of 0.571, with a poisonous and edible species confusion rate of 5.31\% and 2.05\% respectively \cite{2023winner}. 
To handle unknown classes, the team also introduced an entropy based approach to identify unknown, out-of-distribution species \cite{entropic-based-approach}. 

Beyond FungiCLEF, Wei et al. \cite{Wei01} provides a comprehensive examination of fine-grained visual categorization (FGVC) challenges, such as accurately localizing object parts, selecting informative features under varied conditions, and integrating segmentation with classification. 
It emphasizes the need for the model to generalize across species, maintain efficiency, and handle real-world issues like occlusions. 
Other directions that demonstrated promise on FGVC datasets such as CUB-200-2011 \cite{clubdataset} include Mask-CNN \cite{MaskCNN} which outperformed other methods by better capturing subtle differences between species, and SR-GNNs \cite{SR-GNN} which extracted context-aware features from relevant image regions to discriminate between object classes. 

\section{Methodology}

Our overall approach to this challenge of fine-grained classifying of fungi species was to:

\begin{enumerate}
\item Incorporate metadata as additional input / prediction targets for the model.
\item Learn the concept of unknown classes by incorporating the validation dataset into training.
\item Experiment with objective functions to induce model capability in fine-grained classification task.
\item Train only on metadata and image embeddings for rapid prototyping and model optimization.
\end{enumerate}


Cloud computing resources were supplied by \textit{Data Science @ Georgia Tech}. Data was hosted on Google Cloud Storage, and models developed on virtual instances with NVIDIA L4. For more memory intensive experiments, GPUs used in model development include NVIDIA RTX 4090 and a distributed cluster with 2x NVIDIA V100. 


Libraries used include pandas \cite{Pandas}, PaCMAP \cite{PaCMAP}, scikit-learn \cite{scikit-learn} for data exploration; PySpark \cite{PySpark}, PyArrow \cite{PyArrow}, Luigi \cite{Luigi} for data processing; PyTorch \cite{PyTorch}, timm \cite{timm}, Lightning \cite{Lightning}, and transformers \cite{Transformers} for model development. Evaluation functions for the FungiCLEF competition were referenced in the development of internal model benchmarks \cite{DanishFungiGithub}. 

\begin{figure}[b]
  \centering
  \includegraphics[width=.65\textwidth]{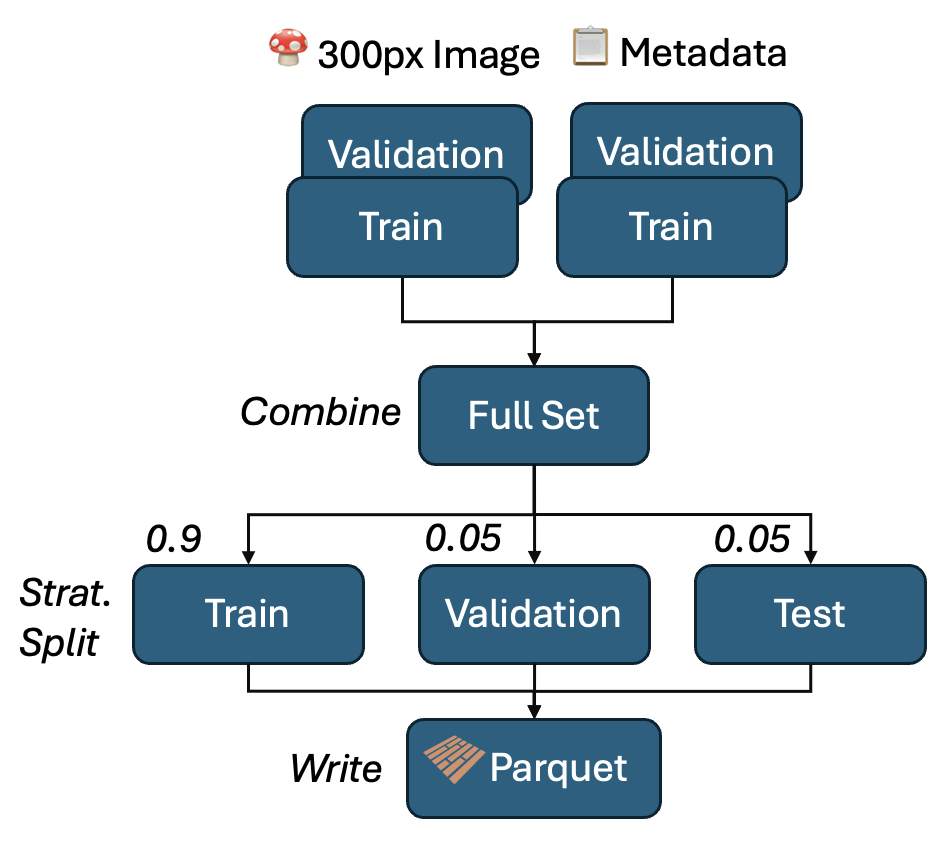}
  \caption{Dataset preparation pipeline: In order to include the unknown classes of fungi from the validation dataset into model training, we mixed the training and validation datasets into a full dataset for further dataset splitting.}
  \label{fig:pre-process}
\end{figure}

\subsection{Dataset Preparation}


To improve the efficiency of experiments, we built a data preprocessing pipeline with PySpark (Figure \ref{fig:pre-process}). We appended the 300 pixel and full versions of image data with associated metadata, and stored them as parquet files for faster I/O. Embeddings were also precomputed and stored as parquet files separately. A custom PyTorch dataset object was created to serve the image and embedding data alongside metadata.


The metadata columns were grouped based on their potential use as either model inputs or prediction targets. For the validation set / public test set, only substrate, metasubstrate, habitat, date, and location were provided \cite{DanishFungiGithub}. As such, these columns were used as additional inputs to the model. Categorical columns were expanded into one-hot vectors. For date information, we converted the month and day into cyclical encoding using sine / cosine transformation \cite{cyclical_encoding}. For location data, we converted longitude and latitude into Geohash, which preserves spatial ordinality and unifying location inputs using a Z-order curve \cite{GeoHash}. Levels 2-5 of the resultant Geohash were extracted and converted from base-32 to normalized base-10 integers. The toxicity and one-hot vectors of the taxonomical levels of fungi classes were included as additional prediction targets. Other metadata columns were excluded.


Given that unknown classes were only present in the validation dataset, we divided DF21 into three equal sections of 20,000 cases, stratified by species. One of the sections was designated as the held out test set, with the remaining two sections utilized as validation set / addition to the training set in a two-fold cross validation training. By percentage, this gives us a ratio of 90.4\%, 4.8\%, 4.8\% for training, validation, and testing over the entire dataset. Significantly, the test and validation sets in each validation fold have the same class distribution.

\subsection{Embeddings for Transfer Learning}

Embeddings are the learned intermediate representation of deep learning models that capture structure about the input domain.
We experimented with two models as the vision model backbone to generate embeddings - DINOv2 \cite{oquab2024dinov2} and ResNet \cite{He01}. DINOv2 was chosen as it is state-of-the-art in terms of vision model and its richness and robustness as a visual feature extractor \cite{oquab2024dinov2}. ResNet was chosen due to its widespread application and downstream \cite{He01}, and serves as a representative for the CNN family in contrast to the transformer family where DINOv2 originated from. For ResNet18, we generated embeddings by extracting the output features from the last hidden state before the classification head. This resulted in embeddings of shape (1000, ) per image. For DINOv2, we utilized the [CLS] token from the last hidden state of model output. In initial experiments and ablation studies, \textit{dinov2-small} \cite{oquab2024dinov2} was used, which results in embedding shape of (768, ). In our optimized model used for competition submission, \textit{dinov2-large} with register \cite{oquab2024dinov2} was used, which had embedding shape of (1024, ). 

For training, the image embeddings were precomputed. For the testing set and for our competition submission, the vision backbone model was frozen, and embeddings were generated during inference and fed into our trained classifier heads.

\subsection{Model Development}

We explored two separate approaches in model development (1) Training a computer vision model from end-to-end, and (2) Training a classifier head only on precomputed embeddings. While approach (1) is the more traditional method for computer vision tasks, it is much more compute intensive due to the number of parameters to be trained \cite{CV-Training}. In comparison, approach (2) had significantly less memory requirements and faster training time (Table \ref{tab:training-specifications}). While using precomputed image embeddings imply that training data could not undergo traditional augmentation techniques in computer vision such as flipping and random cropping, we hypothesize that modern vision models had sufficient amount of information in the feature representation that the downstream model can be robust and generalisable. 

\begin{table}[ht]
  \caption{
  Memory and compute requirements for a fine-tuned vision backbone model (represented by MetaFormer) v.s. transfer learning model (our approach). 
  Overall, the transfer learning model had less memory overhead and faster training time.
  }
  \label{tab:training-specifications}
  \begin{tabular}{ccccl}
    \toprule
    & Trainable & Batch Size & Training Time & VRAM\\
    & Parameters & & per Epoch (s) & Requirement\\ 
    \midrule
    Vision Backbone Model (MetaFormer) & 69M & 64 & 2580 & 60GB\\ 
    Transfer Learning Model (DinoV2) & 47M & 512 & 67 & 8GB\\
    \bottomrule
  \end{tabular}
\end{table}

\subsubsection{Model Training}
For transfer learning with the embedding model, we use a traditional MLP classifier head with a hidden dimension of 4096, with metadata directly concatenated to the embedding. 
Inspired by Diao et al. \cite{Diao}, we also experiment with using a transformer block for better integration of metadata into the classifier. 
We transformed metadata into the same dimensions as the embedding with a separate MLP layer, and added them to the image embeddings before streaming all the data into a transformer block for image classification. 
To leverage the benefits of cross fold validation, we utilized an ensemble model approach \cite{ensemble}. Output logits of our model are averaged over all the classifier heads. 


All models were first trained on a smaller, exploratory development set, before undergoing training runs on the full dataset. For experiments that appeared promising in its initial outcomes, training parameters were further tuned using Optuna to generate a full model for benchmark. Training performance were logged on Weights \& Biases, with the top 2 performing models saved as checkpoints. A two-fold cross-validation was used, where each fold had 1/3rd of DF21 dataset as the validation set, and another 1/3rd incorporated into the training data. Our experiment logs can be viewed at \url{https://wandb.ai/chiu/FungiClef}. 


All experiments were trained 20 to 50 epochs each, with batch sizes of 64 to 512. 
Initial learning rates ranged from $1\cdot10^{-5}$ to $1\cdot10^{-3}$, with AdamW \cite{AdamW} as optimizer. 
Learning rate schedulers experimented with include cosine scheduler with restarts \cite{WarmRestart}, and ReduceLROnPlateau \cite{ReduceLROnPlateau}. 


Metrics recorded during training include training / validation loss, top-1, top-3 accuracy, macro F1 score, and accuracy for correct identification of poisonous species. Calculation for specific track scores were adapted from the FungiCLEF competition \cite{fungiclef2024} for model benchmark. This includes classification error (Track 1), cost for poisonousness confusion (Track 2), and user specific cost (Track 3) \cite{DanishFungiGithub}.

\subsubsection{Loss Function}

The baseline loss function for model development was unweighted, multi-class cross entropy loss. We also explored incorporating class weights in cross-entropy loss, and other loss functions such as focal loss \cite{FocalLoss} and seesaw loss \cite{SeesawLoss}, which was used by last year's winner \cite{2023winner} to overcome class imbalance. Additionally, we experimented with using various metadata such as the higher level taxonomy of the fungi class and the toxicity of the fungi class as additional prediction targets.

For our benchmark model, the model was trained with a custom loss function: \[L_{\text{composite}} = L_{\text{seesaw}} + \alpha \cdot L_{\text{poison}}\]

Where \( L_{\text{seesaw}}\) is the seesaw loss of the class prediction, \( L_{\text{poison}}\) is the binary cross entropy loss of the model's prediction in whether the fungi is poisonous, and \(\alpha\) an adjustable weighting factor for the composite loss function. 
 
\subsubsection{Weighted Sampling}

While weighted sampler is usually utilized to overcome class imbalance \cite{class-imbalance}, we utilized this technique in our data loader to ameliorate the difference in class distribution between the training and validation set. Instead of adjusting class weights such that each class is evenly represented, we derived the per-sample weight by dividing the class frequency of the validation set over the training set:
\[W_{sampling} = \frac{\mathcal{D}_{training}}{\mathcal{D}_{validation}}\]

\section{Results}

\subsection{Training Results}

Our best performing model was an ensemble model on DINOv2 embeddings consisting of two classifier heads (180MB each) from the two-folds of cross-validation training. 
The model was trained on image embeddings precomputed from DINOv2-large. 
The weighting for poison loss \(\alpha\) was 0.1. The initial learning rate was \(1\cdot10^{-4}\), with AdamW \cite{AdamW} as optimizer, and cosine learning rate scheduler with warm restarts \cite{WarmRestart}. 

\begin{table}[ht]
  \caption{Performance of the top 3 models from each training fold and ensemble model across various FungiCLEF metrics.}
  \label{tab:ensemble_benchmark}
  \begin{tabular}{cccccl}
    \toprule
    Fold & Acc. & Track 1 & Track 2 & Track 3 & F1\\
    \midrule
    1 & 74.9\% & 0.251 & 0.414 & 0.665 & 0.409\\
    1 & 74.2\% & 0.258 & 0.418 & 0.675 & 0.406\\
    1 & 74.1\% & 0.259 & 0.430 & 0.689 & 0.378\\
    2 & 73.0\% & 0.270 & 0.287 & \textbf{0.557} & 0.374\\
    2 & 72.2\% & 0.278 & \textbf{0.282} & 0.560 & 0.405\\
    2 & 75.0\% & 0.250 & 0.322 & 0.571 & 0.391\\
    Ensemble & \textbf{78.0\%} & \textbf{0.221} & 0.385 & 0.606 & \textbf{0.451}\\
    \bottomrule
  \end{tabular}
\end{table}

\begin{table}[ht]
  \caption{Ablation study of DINOv2 embedding classifier with various configurations of class weighting and incorporation of metadata as input or additional prediction targets.}
  \label{tab:DINOv2_ablation}
  \begin{tabular}{ccl}
    \toprule
    &Val. Acc. (\%)&Difference (\%)\\
    \midrule
    DINOv2 baseline & 49.3 & -\\
    w/ class weighting & 65.5 & +16.2\\
    w/ metadata & 69.1 & +3.6\\
    incl. toxicity & 68.9 & -0.1\\
    incl. taxonomy & 69.5 & +0.5\\
    w/ weighted taxonomy & 69.7 & +0.2\\
    \bottomrule
  \end{tabular}
\end{table}


Table \ref{tab:ensemble_benchmark} outlined the three top performing models from each fold. Overall, Fold 1 had worse performance on the held out test set than Fold 2. The ensemble model with the best performing model from each fold yielded better accuracy than either folds, but suffered partially with poisonous performance. We also experimented with incorporating more models in the ensemble model, but the model performance was worse, likely due to the model negatively biased by inferior models. 

To explore the impact of various factors in model performance, we performed an ablation study with results summarized in Table \ref{tab:DINOv2_ablation}.
The ablation experiments were done using smaller embeddings and a MLP head instead of a transformer head.
We find that class-weighting increased the scores by the largest margin, followed by the inclusion of metadata.

\subsection{Leaderboard Results}

The results of our team's experiments are outlined in Table \ref{tab:final_results}. For our first submission during competition, we used a pre-trained MetaFormer model from the previous year's competition as a baseline. In post-competition evaluation, our best model achieved an accuracy of 78.4\% and a macro F1 score of 0.577 in the private test set. 
Our model's performance was comparable to previous years' winners \cite{2023winner}, and was the best performing model in this year's competition in terms of Track 1, Track 3, and accuracy.
Our Track 2 and F1 score was ranked 2nd compared to the rest of the competitors \footnote{Due to numerous issues with the HuggingFace platform, our best results were not recorded in the official competition. Our post competition evaluation was performed under the same constraints as the official competition. Post-competition results were provided and verified by the organiser of FungiCLEF.}. 
The inference time across the full public test set (40,216 images) was 25:26 minutes, and 0.126s per image on average on a RTX 4090.

\begin{table}[ht]
  \caption{Public and private test set scores on the official leaderboard and post-competition evaluation.\protect\footnotemark.
  }
  \label{tab:final_results}
  \begin{tabular}{llrrrrr}
    \toprule
    & Name & Track 1 $\downarrow$ & Track 2 $\downarrow$ & Track 3 $\downarrow$ & F1 $\uparrow$ & Acc. $\uparrow$ \\
    \midrule
    Private & MetaFormer (Competition) & 0.391 & 1.604	& 2.044	& 30.0 & 60.9 \\
    Private & DINOv2 (Post Competition) & 0.216 & 0.129& 0.345 & 57.7 & 78.4 \\
    \midrule
    Public & MetaFormer (Competition) & 0.395 &  1.649 & 2.044 &  27.6	&  60.5	\\
    Public & DINOv2 (Post Competition) & \textbf{0.211} & 0.165& 0.375 & 49.8 & \textbf{79.0} \\
    \midrule
    Public & Rank 1 - IES 	& 0.2922 	& \textbf{0.0699} 	& \textbf{0.3621} 	& \textbf{54.99} 	& 70.78 \\
    Public & Rank 2 - jack-etheredge 	& 0.2394 	& 0.1681 	& 0.4075 	& 49.81 & 76.06 \\
    Public & Rank 6 - Baseline with EfficientNet-B1 	& 0.4926 	& 0.6599 	& 1.1526 	& 32.99 & 50.74\\
    \bottomrule
    \end{tabular}
\end{table}

\footnotetext{
Official competition results are from test submissions with an under-tuned vision model. 
These results are included for completeness.
}

\section{Discussion}

We intially experimented with vision models including EfficientNet \cite{EfficientNet}, VisionTransformer \cite{ViT}, and MetaFormer \cite{MetaFormer}. 
Due to training time and memory overhead, we opted to focus our efforts into developing a lightweight classifier on precomputed embeddings instead.

\subsection{ResNet v.s. DINOv2 as Vision Backbone for Embedding Generation}

\begin{figure*}[ht]
  \begin{center}
    \centering
    \subfloat{%
        \includegraphics[width=0.75\textwidth]{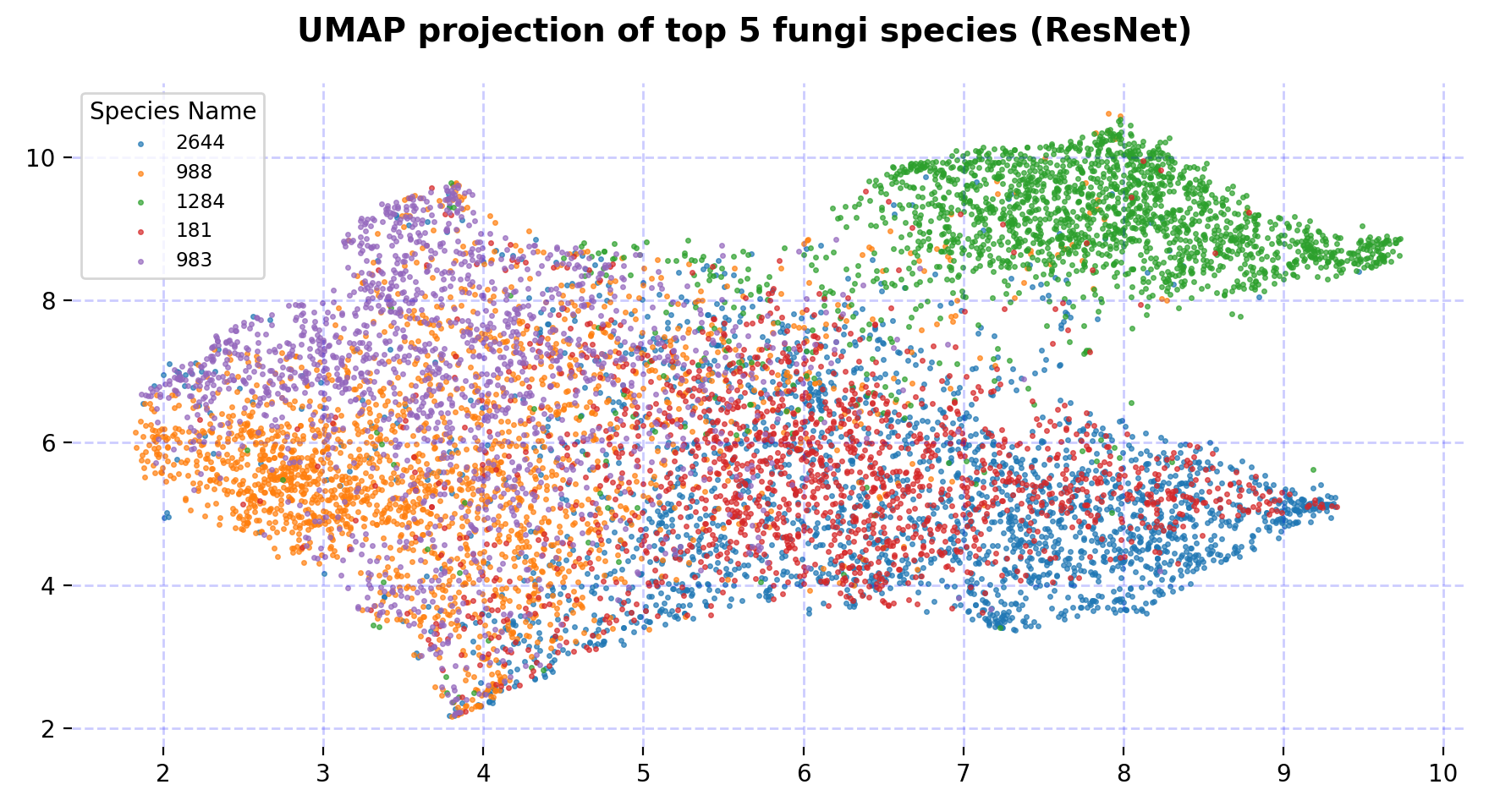}%
        \label{fig:resnetumap}%
        }%
    \hfill%
    \subfloat{%
        \includegraphics[width=0.75\textwidth]{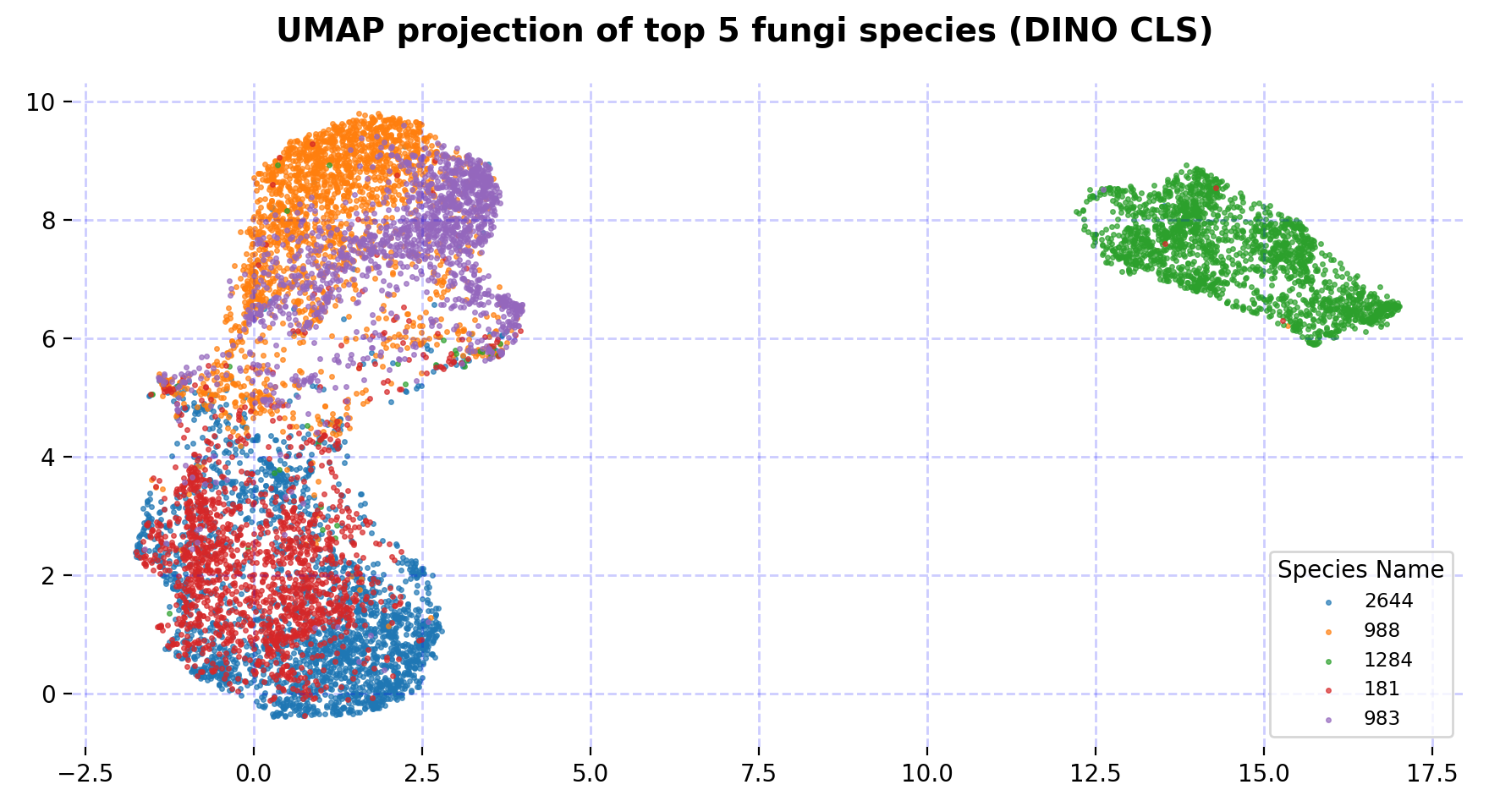}%
        \label{fig:dinoumap}%
        }%
    \hfill%
  \end{center}
  \caption{
  Clustering of top 5 fungi species on ResNet and DINOv2 with UMAP. 
  We observe that ResNet embeddings do not separate well, but there were clear separation of clusters in DINOv2.
  }
  \label{fig:umap}
\end{figure*}

Overall, while DINOv2 embeddings proved to be good input for image classification, our embedding model using ResNet embeddings did not perform well, with best validation accuracy at 25\%. 
This was likely due to DINOv2 being a class agnostic, self-supervised model, whereas ResNet was trained on ImageNet with specific classification targets. 
As such, the features extracted from ResNet would be more tailored to the dataset, whereas DINOv2 features were more representative of the underlying image \cite{oquab2024dinov2}. 
To further investigate this, we visualized the embeddings with UMAP \cite{UMAP} in Figure \ref{fig:umap}, which showed that ResNET embeddings did not separate well, whereas there was a clear separation in DINOv2 embeddings. 

\subsection{Incorporation of Metadata}

We experimented with using metadata as additional prediction targets as seen in our ablation in Table \ref{tab:DINOv2_ablation}.
However, this did not yield additional performance, but the additional overhead required to tune the weighting of various targets was not worth the complexity. 
As such, we did not utilize metadata in our final model. 
The inclusion of metadata appeared to provide some marginal benefits in validation accuracy and F1 score. 
This echoes the finding from previous research on the dataset, where the incorporation of metadata as input had a positive contribution to model performance. 

\section{Future Work}

Whilst using embeddings allowed for much faster model development time, there is still an additional gap in the performance of the embedding classifier compared to traditional image-based models. It is likely that the information loss in the transformation of image to embeddings was too significant for the simple classifier architecture to overcome. It would be interesting to further fine-tune DINOv2 on the DanishFungi dataset, and repeat our experiments. Moreover, a more rigorous incorporation of metadata into our models could provide a more holistic understanding of the data, leading to more accurate and reliable classification systems.

\section{Conclusion}

In summary, we addressed the complex task of fine-grained visual categorization (FGVC) for identifying poisonous fungi using transfer learning and advanced deep learning methodologies. The Danish Fungi 2020 dataset presented significant challenges such as class imbalance, subtle inter-class variations, and high intra-class variability, necessitating a comprehensive data preprocessing and augmentation pipeline.

Our experiments with various deep learning models, including vision transformers, convolutional neural networks, and linear classifiers with embeddings, highlighted the potential of DINOv2 embeddings combined with a multi-layer perceptron. Integrating multimodal metadata further enhanced classification performance, emphasizing the value of auxiliary information. Despite promising results, embedding-based classifiers faced limitations due to potential information loss, suggesting the need for fine-tuning self-supervised models on domain-specific datasets and improved metadata incorporation. Overall, our research advances FGVC technical capabilities, providing valuable methodologies for mycological safety and educational applications, and contributes to the broader field of fine-grained classification tasks.

\section*{Acknowledgements}

We thank the DS@GT CLEF team for providing the development and research environment for our machine learning experiments as well as valuable comments and suggestions.

\bibliography{main}


\end{document}